\documentclass[letterpaper, 10 pt, conference]{ieeeconf}  

\IEEEoverridecommandlockouts                              

\usepackage{times}
\usepackage{epsfig}
\usepackage{graphicx}
\usepackage{amsmath}
\usepackage{amssymb}
\usepackage{soul}
\usepackage[utf8]{inputenc} 
\usepackage[disable]{todonotes}	
\usepackage{color}	
\usepackage{adjustbox}
\usepackage{multirow}
\usepackage{rotating}
\usepackage{tikz}
\def\checkmark{\tikz\fill[scale=0.4](0,.35) -- (.25,0) -- (1,.7) -- (.25,.15) -- cycle;} 

\usepackage[caption=false,font=footnotesize]{subfig}
\newcommand{\citep}{\cite}

\newcommand{\etal}[1]{\textit{et al. }}
\newcommand{\fig}[1]{Fig.~\ref{#1}}

\title{\LARGE \bf
Boosting Supervised Learning Performance with Co-training
}
\author{Xinnan Du, William Zhang, Jose M. Alvarez
\thanks{William Zhang and Jose M. Alvarez are with NVIDIA, USA.   Contact: {\tt josea@nvidia.com}}%
\thanks{Xinnan DU is with Carnegie Mellon University; Work was done while Xinnan was an intern at NVIDIA, USA. 
}
}
\begin{document}

\maketitle
\thispagestyle{empty}
\pagestyle{empty}


\begin{abstract}
Deep learning perception models require a massive amount of labeled training data to achieve good performance. While unlabeled data is easy to acquire, the cost of labeling is prohibitive and could create a tremendous burden on companies or individuals. Recently, self-supervision has emerged as an alternative to leveraging unlabeled data. In this paper, we propose a new light-weight self-supervised learning framework that could boost supervised learning performance with minimum additional computation cost. Here, we introduce a simple and flexible multi-task co-training framework that integrates a self-supervised task into any supervised task. Our approach exploits pretext tasks to incur minimum compute and parameter overheads and minimal disruption to existing training pipelines. We demonstrate the effectiveness of our framework by using two self-supervised tasks, object detection and panoptic segmentation, on different perception models. Our results show that both self-supervised tasks can improve the accuracy of the supervised task and, at the same time, demonstrates strong domain adaption capability when used with additional unlabeled data. 
\end{abstract}
\section{Introduction}
Deep Neural Networks (DNNs) have been extremely successful in recent years by achieving state-of-the-art in almost all computer vision tasks when trained in a supervised manner. DNNs' tremendous representation power has led to their success, but it comes with the price of requiring a massive amount of labeled data. While access to unlabeled data is easy and manageable, the cost of labeling data is usually high, especially for complicated perception tasks such as object detection or panoptic segmentation.  Self-supervised learning has recently emerged to leverage unlabeled data and boost the performance of supervised approaches. The key idea behind self-supervision is to generate automatic labels from data and use those for training the DNNs.

There are different formulations for self-supervised learning. These include pretext-tasks, contrastive learning, and generative modeling. Pretext-tasks transform input data using, for instance, rotations~\cite{gidaris2018unsupervised}, moving patches (jigsaw)~\cite{noroozi2016unsupervised}, or colorizations, and the goal is to recover the original image. Contrastive learning focuses on formulating the task of finding similar and dissimilar images/patches to train a deep network, and generative modeling techniques aim at generating parts of the input data given some contextual information. An example of these generative modeling approaches is in-painting, where we remove a patch of the input image, and the goal is to generate the patch as accurately as possible. Generative adversarial  networks \cite{radford2015unsupervised} \cite{donahue2016adversarial} could also be used for self-supervised learning, where feature learned by generator and discriminator could be used for supervised tasks. In general, using self-supervision for pre-training models on large amounts of data can successfully transfer knowledge to other downstream tasks and outperform the supervised pre-training counterparts. In this work, we focus on the first group, pretext-tasks, as they incur less disruption into existing training pipelines.

In this work, we present our approach to leverage self-supervision to bootstrap the performance of existing supervised learning perception tasks. In particular, instead of pre-training models, we focus on a multi-task learning setting where we train all tasks of interest together, hoping each task will benefit from each other and improve the overall performance. To this end, we introduce a co-training approach where self-training is considered an auxiliary task to help during the optimization process. Our approach exploits classification-based pretext-tasks such as rotation and jigsaw, where the goal is to predict the correct transformation applied to the input image. Therefore, the parameter overhead due to the auxiliary task is minimal, forcing the auxiliary task to rely mostly on the same parameters as the primary task. Intuitively, sharing most parameters between the primary and the auxiliary task will improve the shared representation learning. Besides, this multi-task setting should help to regularize the learning process and reduce over-fitting. 

We demonstrate the effectiveness of our approach on object detection and panoptic segmentation on large datasets. Our experiments demonstrate that our co-training approach consistently outperforms the single task counterpart on both object detection and panoptic segmentation. In panoptic segmentation, we obtain up to 2.3\% relative improvement in overall performance using the same amount of data with a marginal increase in computational cost at train time. Furthermore, our approach lets us leveraging additional unlabeled data that yields to networks that are more robust to domain shifts at inference.

We start summarizing related works for self-supervised and multi-task learning in Sect.~\ref{sect:sota}. Then, in Sect.~\ref{sect:method} we present our co-training approach and, in Sect.~\ref{sect:experiments}, we present our experiments and discuss our results.

\begin{figure*}[!t]
  \centering
  \resizebox{\textwidth}{!}{
  \begin{tabular}{cc}
  \includegraphics[width=0.45\textwidth]{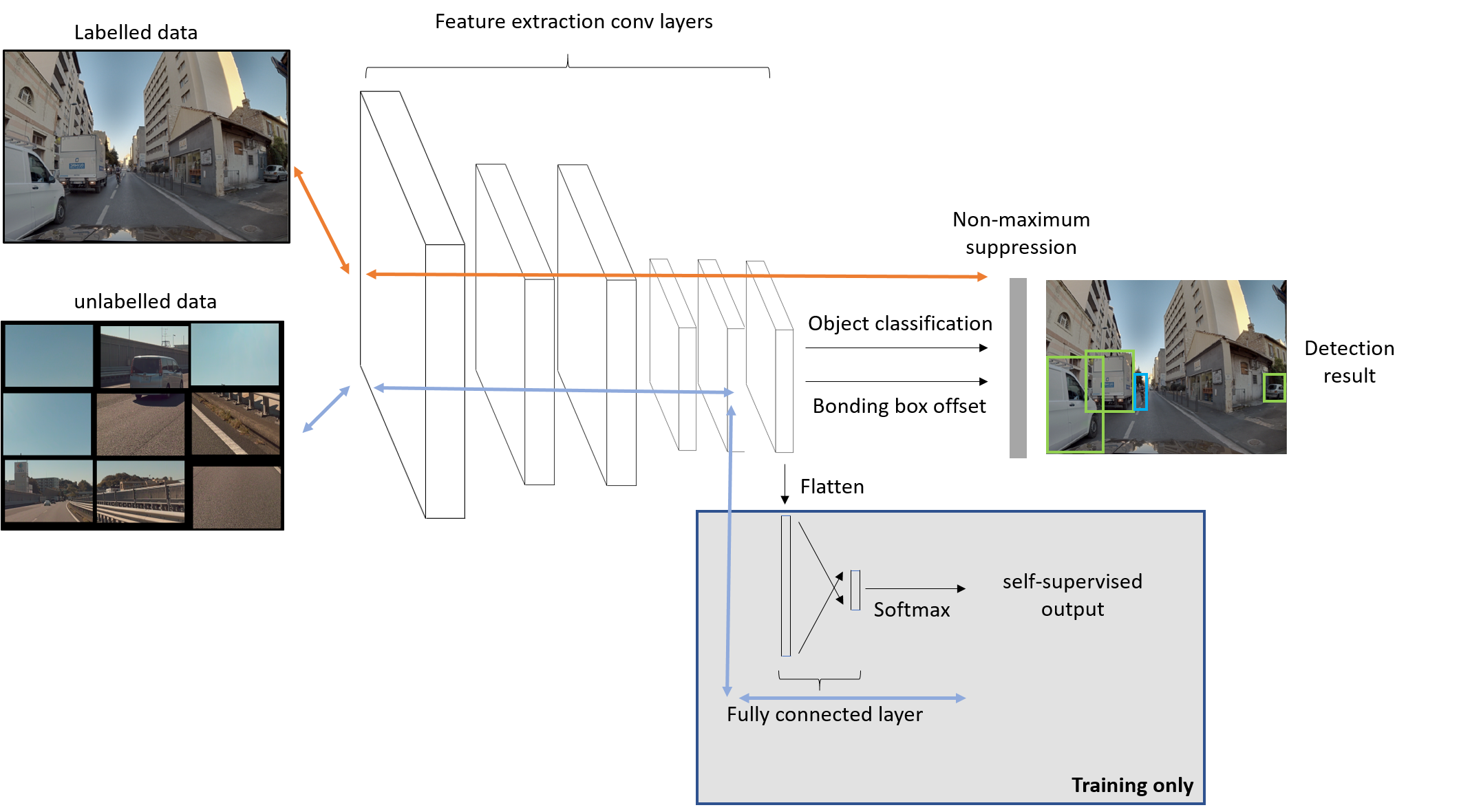}&
  \includegraphics[width=0.45\textwidth]{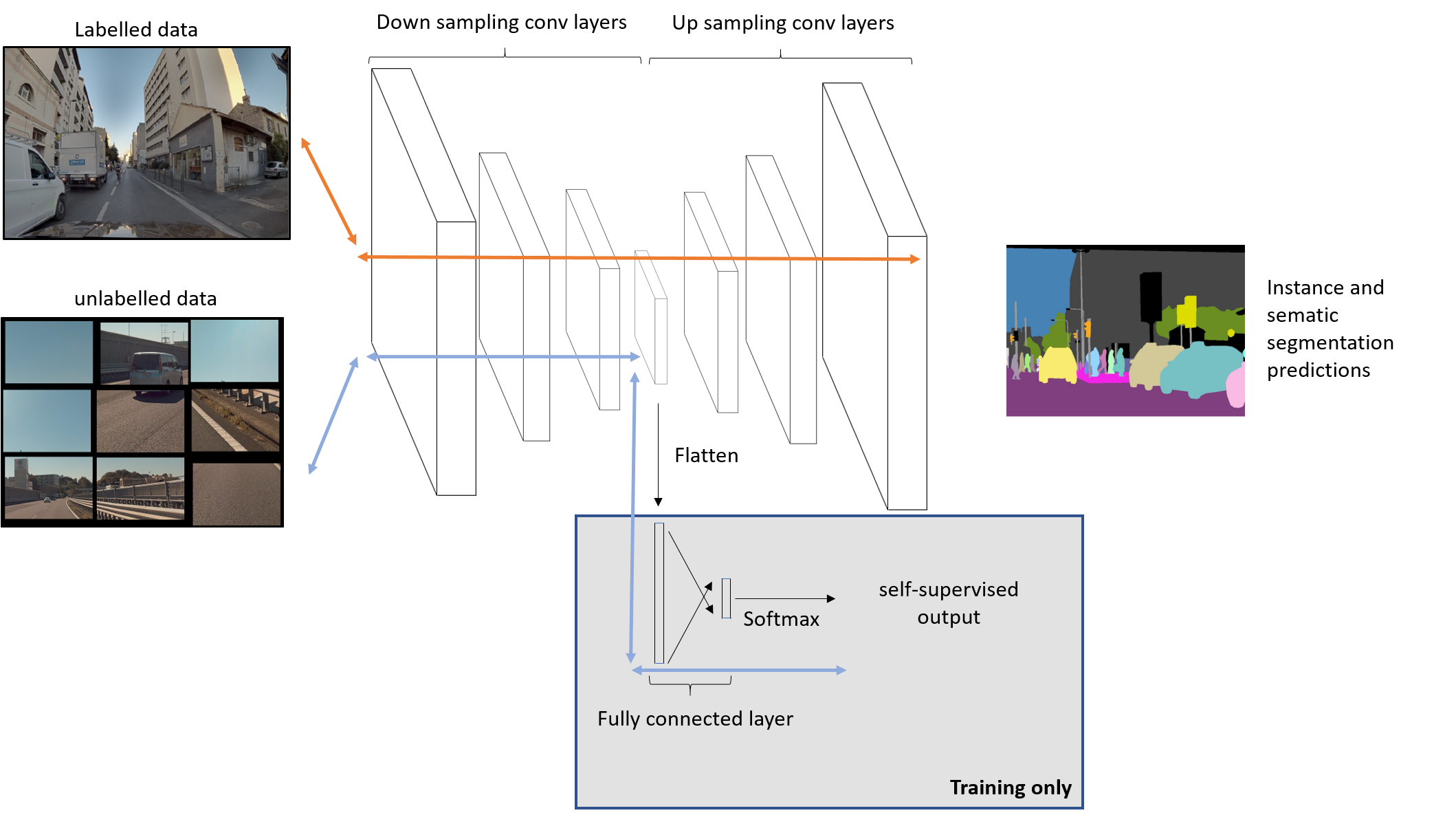}\\
  (a)&(b)\\  
  \end{tabular}
  }
  \caption{Illustration of our architecture for a) object detection and b) panoptic segmentation. For object detection, the self-supervised task branches out at the end of the feature extractor. For panoptic segmentation, the self-supervised branch is added at the end of down sampling layers. Areas in the gray boxes are considered only during training. The orange line shows the data flow of the supervised task and the blue line shows the self-supervised data flow.}
  \label{fig:architecture}
  \vspace{-0.5cm}
\end{figure*} 

\section{Related Work}
\label{sect:sota}
\subsubsection{Self-supervised learning}
Self-supervised learning is a type of unsupervised learning that uses labels that are available to the data without any manual labeling.  Predicting image transformation is a popular way to do self-supervised learning: a set of predefined transformations is randomly applied to images and the network needs to learn to discriminate which type of transformation is applied. These methods are usually cast as classification problems, such as distortion \cite{dosovitskiy2015discriminative}, rotation \cite{gidaris2018unsupervised}, and jigsaw \cite{carlucci2019domain}. Patch-based methods \cite{noroozi2016unsupervised} \cite{doersch2015unsupervised} need the network to reason about the relative positions of patches in images, and hence provide spatial correlation clues to the network. Generative methods are also used as self-supervised pretraining tasks: these include the colorization of images \cite{zhang2016colorful}, denoising auto-encoder \cite{vincent2008extracting}, image inpainting \cite{pathak2016context}, and split-brain auto-encoders \cite{zhang2017split}. Generative adversarial networks \cite{radford2015unsupervised} \cite{donahue2016adversarial} are another way to perform self-supervised learning, as the training process doesn't need any manually labeled data. Another family of self-supervised algorithms is contrast learning which works by encouraging the representation of the original image and transformed image to be similar \cite{chen2020simple} \cite{misra2020self}. The design of such image transformations is crucial to self-supervised performance.

\subsubsection{Multi-task learning}
Multi-task learning has been a popular research topic for years, early works such as \cite{caruana1997multitask} \cite{thrun1998learning} train related tasks jointly with shared parameters and observe that tasks could benefit from each other and reduce overfitting. Entering the deep learning era, many efforts have been devoted to multi-task learning for computer vision. Perception models are usually built with convolutional neural networks, when trained in a multi-task setting, different tasks usually share part of the convolution layers and also have their own task specific parameters~\cite{bilen2016integrated}\cite{misra2016cross}. \cite{long2015learning} used this structure and improves it with matrix priors which allows the network to learn the relationship between tasks. \cite{lu2017fully} proposed a widening network that could improve adaptive feature sharing. Multi-task training could be tricky as tasks may compete with each other. As such, some methods focused on improving the training process: \cite{yang2016trace} regularizes the network with tensor trace norm to encourage parameter sharing, and a sharing strategy is learned rather than predefined. \cite{sener2018multi} casts this problem into a multi-objective optimization to get a Pareto optimal solution.

\subsubsection{Self-supervised learning with multi-task learning}
There are more and more self-supervised learning methods that adopt multi-task learning as a way to improve training.  Some train multiple pretext self-supervised tasks in a multi-task setting \cite{ren2018cross} \cite{doersch2017multi}, which results in superior performance compared to single task self-supervised learning. \cite{han2020self} uses contrastive learning co-training to improve the sampling process so that the network could learn a better video representation. Other approaches combine self-supervised learning and supervised learning and optimize them together in a multi-task setting, \cite{carlucci2019domain} \cite{kundu2019adapt} combines a self-supervised task and a supervised task to improve domain generalization, \cite{klingner2020improved} uses synthetic image depth prediction to increase the robustness of semantic segmentation, and \cite{lee2019multi} trains object detection with multiple self-supervised tasks to improve performance.

\section{Boosting  supervised  learning  performance  with  self-supervised co-training}
\label{sect:method}
In this section, we present our pipeline to boost the performance of existing perception models. To this end, we jointly train the primary task with a self-supervised auxiliary task in a co-training fashion. Instead of pre-training the model using self-supervision and then fine-tuning the model, we directly optimize both tasks simultaneously. \fig{fig:architecture} shows an overall view of our approach. Intuitively, both tasks share the same feature extractor, and therefore, the self-supervised branch would improve the feature extractor by providing extra latent information about the training images. 

Let's assume the supervised training dataset contains \(N_1\) datapoints \(\{(x_i, y_i)\}_{i=1}^{N_1}\) where \( x_i \) are labeled images and \( y_i \) are the corresponding labels. On the self-supervised branch, we assume \(N_2\) raw images \(\{{a_i}\}_{i=1}^{N_2}\), then, the self-supervised algorithm will apply some processing \(g\) on the image to auto-generate self-supervised training data together with labels and obtain  \(\{({g(a_i)}, {b_i)}\}_{i=1}^{N_2}\). For the supervised branch, the output of the model is given by
\begin{equation}
{\hat{y}_i} = h(x_i\mid\theta_a, \theta_b),
\end{equation}
\noindent where \(\theta_a\) refers to the parameters of the feature extractor and \(\theta_b\) refers to parameters specific for the supervised task. We optimize these parameters by minimizing a supervised task loss 
\begin{equation}
L_{sup} = \sum_{i=1}^{N} \ell_{sup}(h(x_i \mid\theta_a, \theta_b), y_i),
\end{equation}
\noindent where \(\ell_{sup}\) refers to the loss for the task (e.g., softmax for classification). Similarly, for the self-supervised task, the output of the model is given by
\begin{equation}
{b'} = h(g(x_i)\mid\theta_a, \theta_c),
\end{equation}
\noindent where $g(.)$ refers to a transformation to the input image and \(\theta_c\) refers to the task specific parameters for the self-supervision branch. We optimize the self-supervision branch by solving minimizing the self-supervised loss,
\begin{equation}
L_{self} = \ell_{self}(h(g(x_i)\mid\theta_a, \theta_c), b),
\end{equation}
\noindent where $b$ is the automatically generated labels for the self-supervised task.

During training, we minimize the combined loss of both supervised and self-supervised branches
\begin{equation}
L = L_{sup} + \omega L_{self},
\end{equation}
where \(\omega\) is the weight coefficient of the self-supervised task. We train these tasks in an alternating fashion which allows us to control the learning pace of the two tasks and give the framework additional flexibility. 

Let R ($R>>1$) be the training ratio between the two tasks. At each training step, the trainer will select the supervised task with probability \(R/(R+1)\) and select the self-supervised task with probability \(1/(R+1)\). In this way, the compute overhead is \(1/R\) and becomes negligible when R is large.


\begin{figure}[!t]
  \centering
  \includegraphics[scale=0.225]{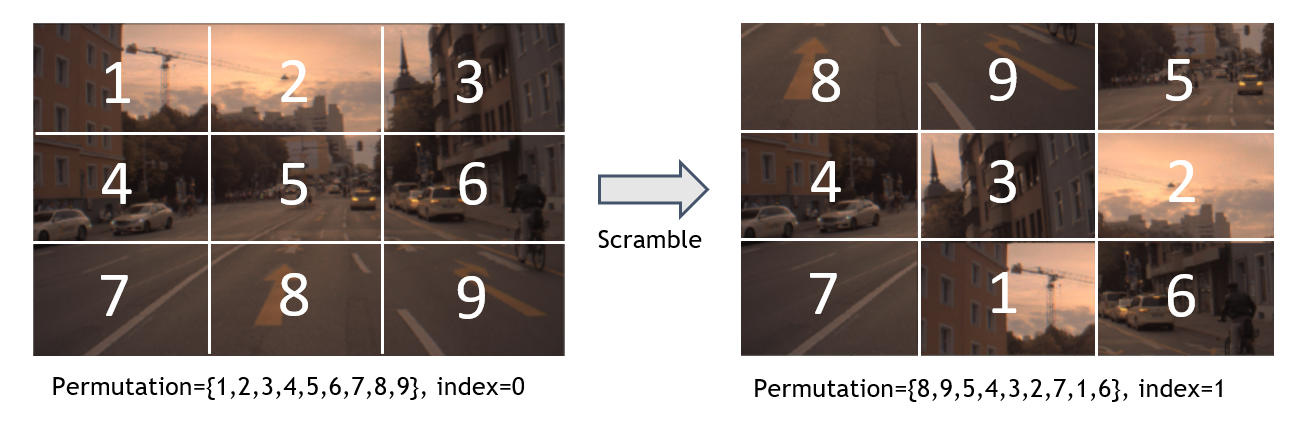}
  \vspace{-0.5cm}
  \caption{\textbf{Jigsaw Task.} The input image is scrambled according to a permutation and the goal is to predict the permutation applied.}
  \label{scramble}
\end{figure} 
\begin{figure}[!t]
  \centering
  \includegraphics[scale=0.20]{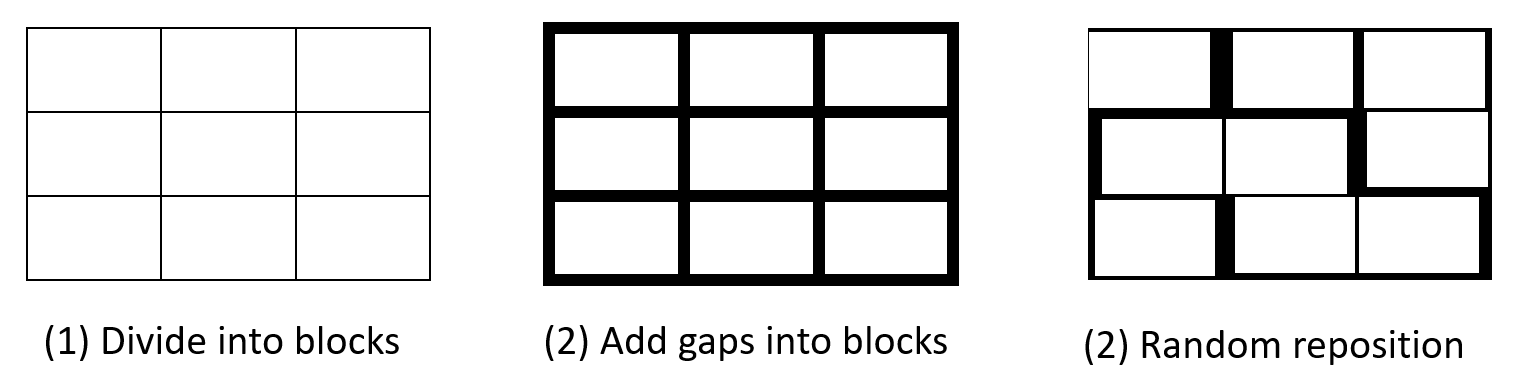}
  \vspace{-0.5cm}
  \caption{\textbf{Jigsaw Task.} We add random gaps between blocks to prevent learning trivial features such as edges between blocks.}
  \label{randomGap}
  \vspace{-0.5cm}
\end{figure} 

\subsection{Self-supervised tasks}
Many self-supervised tasks require complicated reasonsing or reconstruction of an image such as colorization \cite{zhang2016colorful}, image inpainting \cite{pathak2016context}, and Generative adversarial networks~\cite{radford2015unsupervised}. In order to add the minimum amount of extra parameters, we focus on pretext classification tasks which, in practice, only require one extra fully connected layer. In short, these pretext-tasks apply a transformation to the input image and the goal is to predict the right transformation. In our experiments, to showcase the effectiveness of self-supervised co-training, we use two self-supervised tasks: jigsaw and rotation. 
\subsubsection{Jigsaw Task}
The first pretext task we consider is solving Jigsaw puzzles~\cite{noroozi2016unsupervised}. The key idea is to break the input image into patches and scramble those according to a permutation pattern. Then, the goal is to reconstruct the input image from its shuffled parts, see~\fig{scramble}. Intuitively, the jigsaw task could improve the feature extractor by providing useful information about spatial correlation of the images and prevent the other task from overfitting. More precisely, we model the jigsaw task as a classification problem to minimize the amount of extra parameters. To this end, we follow the algorithm proposed in~\cite{carlucci2019domain}, and first divide the image into \(N\times N\) blocks, then we have \(N^2!\) ways to rearrange the \(N\times N\) blocks into a new image. When N=3, we will have \(9!=362880\) permutations which is impractical for classification. Thus, only \(P\) permutations are picked which includes the original image, and the rest of permutations are picked by maximizing Hamming distance described in \cite{noroozi2016unsupervised}. The processing function \(g_{jigsaw}\) scrambles the original image according to the permutation, and the generated label \(b_i\) is the permutation index where \(b_i \in \{0, 1, 2, ..., P-1\}\) as shown in Fig.~\ref{scramble}. 

To prevent the network from learning trivial features such as the edges between the blocks, the processing also include an extra step of adding random gaps between blocks. To add the gap, we first chop the center of the block and leave gaps to edges of the block with width \(G\) and reposition those chopped parts into a random location within their blocks as shown in Fig.~\ref{randomGap}.
\begin{figure}[!t]
  \centering
  \includegraphics[scale=0.20]{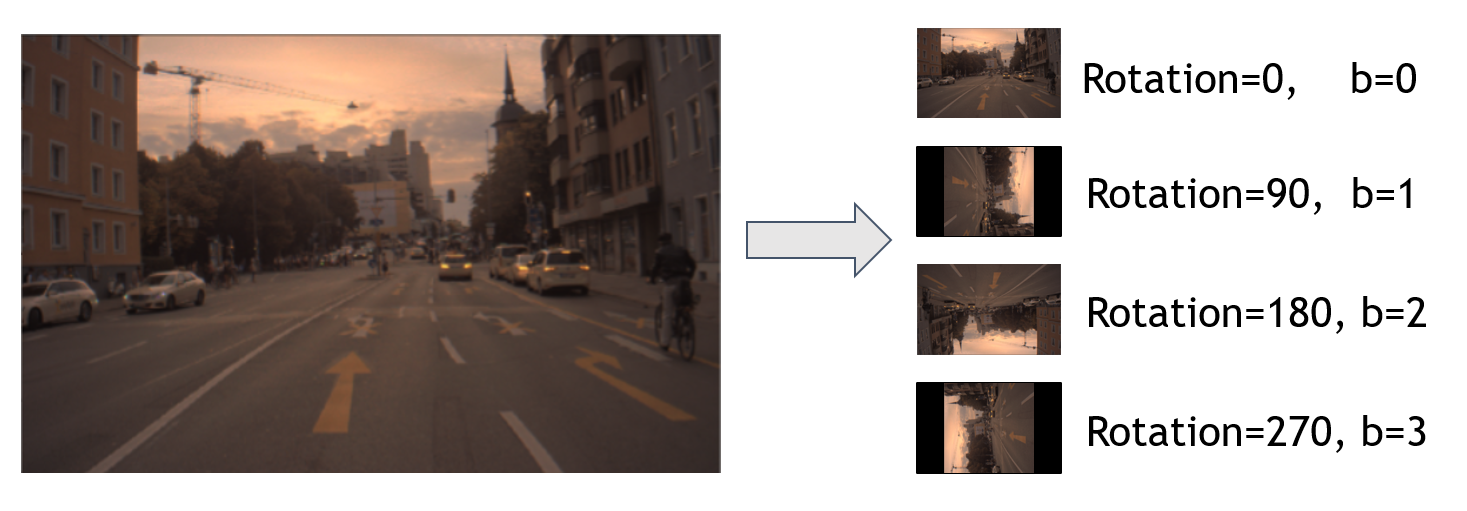}
  \vspace{-0.5cm}
  \caption{\textbf{Rotation task.} The input image is rotated and the goal is to predict the rotation angle.}
  \label{rotation}
  \vspace{-0.5cm}
\end{figure} 
As we formulate the task as a classification problem, the loss function \(\ell_{self}\) is the cross-entropy loss and the training objective of the self-supervised task is
\begin{equation}
    \underset{\theta}{min} \frac{1}{N_2}\sum_{i=1}^{N_2}\ell_{self}(h(g_{jigsaw}(x_i)\mid\theta_a, \theta_c), b_i).
\end{equation}


\subsubsection{Rotation Task}
The second task we consider is Rotation~\cite{gidaris2018unsupervised}. In short, the idea is to rotate the input image and predict the rotation angle, see~\fig{rotation}. Predicting the rotation of an image could guide the network to focus on "important" parts of the image as the objects are the key to recognizing rotations. Similar to the jigsaw task, we also model the rotation task as a classification problem. We start by picking \(K\) rotations and the rotation granularity will then be \(\frac{360}{K}\) degrees. The processing function \(g_{rotation}\) rotates the image by \(i\times\frac{360}{K}\) degrees where \(i \in \{0, 1, ... , K-1\}\) and the label \(b_i\) is the rotation index \(i\). When rotating the image, the center of the image is used as the pivot and out of boundary portions are chopped as shown in \ref{rotation}. The training objective of the rotation task is
\begin{equation}
\underset{\theta}{min} \frac{1}{N_2}\sum_{i=1}^{N_2}\ell_{self}(h(g_{rotation}(x_i)\mid\theta_a, \theta_c), b_i).    
\end{equation}


\section{Experiments}
\label{sect:experiments}
In this section, we provide an exhaustive evaluation of different self-supervised tasks and their benefits on two different computer vision tasks: object detection and panoptic segmentation. 

Below, we first detail the object detection and panoptic segmentation settings, general experimental settings and datasets, and then, we discuss our results. We then also analyze the robustness of our approach to domain shifts and out of distribution samples. 


The goal of object detection is to accurately predict object categories and their corresponding bounding boxes within an image. In the prediction process, each of these predictions would be accompanied with a confidence score, which is a probability score for how likely our algorithm believed each object belongs to a particular categorie. In our experiments, we use a one-stage object detector based on a UNet-backbone. We add the classification head for the self-supervised task at the end of the feature extractor as shown in~\fig{fig:architecture}a. On the other hand, panoptic segmentation combines instance segmentation and semantic segmentation to assign a class label to each pixel and, in addition, detect and segment each object instance independently~\cite{kirillov2019panoptic}. In this case, we use a basic encoder-decoder architecture also based on a UNet-backbone. Here, the classification head of the self-supervised task is added at the end of the encoder, see~\fig{fig:architecture}b. For both, panoptic segmentation and object detection, we empirically found these to be the best performing branching locations.


For the jigsaw task, we use \(N=3\times3\) blocks, \(P=30\) or \(P=100\) permutations, and random gap \(G=20\) pixels. For the rotation task, we use number of rotations \(K=4\). For all experiments, we choose training ratio \(R=6\), self-supervised weight \(\omega=1\), and are trained on 4 Titan V100 GPUs. By default, the self-supervised branch uses the same image as the supervised branch. When comparing with the baseline, the total number of training steps is the same, which means that the supervised task is trained less steps compared to the baseline.

\textbf{Datasets:} In our experiments, we use an internal large scale research datasets. For object detection, the dataset consists of 847k training images and 33K test images. Each image is annotated with 3 classes: vehicle, person, cycle. For evaluation, we report the weighted mean average precision (wMAP) which averages MAP across several object sizes. The dataset for panoptic segmentation consists of 22K training images and 5k test images. Each image is annotated with 4 object classes namely Vehicle, Bicycle, Person, and Road. In this case, we use the IoU metric for evaluation.

\begin{table}[!t]
\caption{\textbf{Panoptic segmentation}.Comparison of our approach to the baseline with different self-supervised tasks for models trained during 250K steps.}
\label{tab:panoptic}
\begin{center}
\begin{tabular}{c c c}
\hline
 Model & self-sup.  task& IoU (\%)\\
 \hline
 baseline & -- &89.07\\  
 Co-train & Jigsaw & 90.14 \\
 Co-train & Rotation & \textbf{90.82} \\
\hline
\end{tabular}
\end{center}
\vspace{-0.2cm}
\end{table}

\subsection{Results}
In this experiment, we first evaluate the effectiveness of our approach to improve the performance of both tasks. We compare the performance of our approach to the baseline. That is, the basic task without co-training. Table~\ref{tab:panoptic} summarizes the results for panoptic segmentation for both jigsaw and rotation tasks. As shown, co-training with self-supervision improves the performance. As we can see, for panoptic segmentation there is a considerable improvement on IoU in both tasks with an improvement up to 2\%, which is significant given the baseline IoU is already near 90\%. Results for object detection are shown in Table~\ref{tab:detection}. In this case, we train the networks for 500k iterations. As shown, the three approaches perform similarly with a slight improvement when using rotation over the baseline and jigsaw. These experiments show that rotation yields even better results than jigsaw despite its simplicity. In any case, we can conclude that co-training with a self-supervision task improves the performance of the baselines.

\begin{figure*}[!h]
  \centering
  \includegraphics[width=0.95\textwidth]{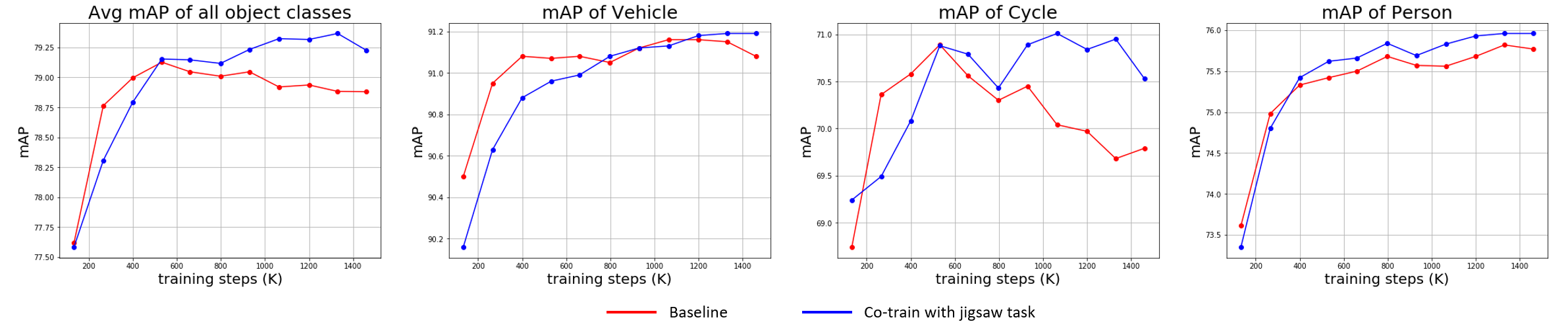}
  \caption{\textbf{Object detection / Jigsaw}. Accuracy as a function of training steps. From left to right, overall average mAP and mAP for each of the classes in the dataset. Our approach, in the long term, consistently outperforms the baseline. The improvement is even larger for underrepresented classes such as cycle. }
  \label{fig:detectionsteps}
  \vspace{-0.5cm}
\end{figure*}

\begin{figure}[!t]
  \centering
  \includegraphics[width=0.6\columnwidth]{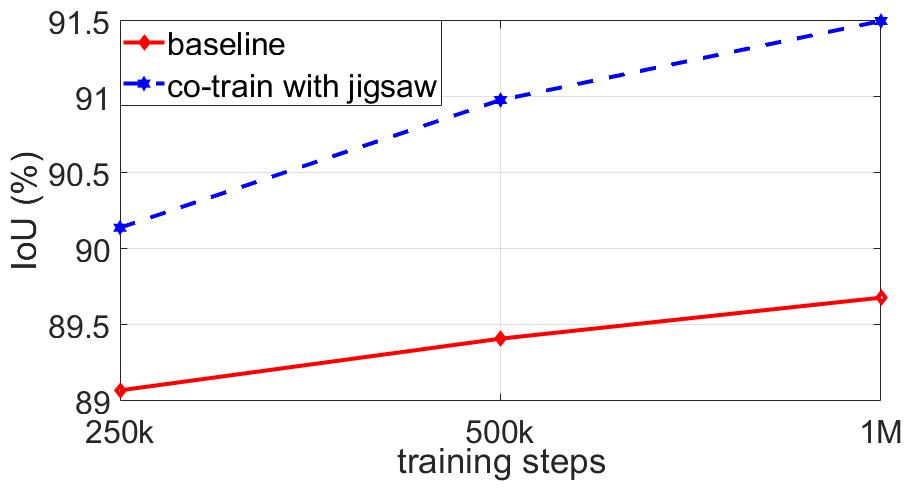}
  \vspace{-0.5cm}
  \caption{\textbf{Panoptic Segmentation / Jigsaw}. Accuracy as a function of the training steps.}
  \label{fig:panopticsteps}
  \vspace{-0.45cm}
\end{figure}

We now take a closer look to these results and analyze the performance of co-training with the jigsaw task as a function of the training iterations. Fig.~\ref{fig:detectionsteps} shows the overall and per-class performance of this experiment for object detection and Fig.~\ref{fig:panopticsteps} shows the results for panoptic segmentation. As shown, in both tasks, longer training yields slight improvements in the overall accuracy as the number of training iterations increases for both the baseline and our approach. Interestingly, for object detection, in long training regimes, co-training is able to outperform the baseline. This behaviour is consistent over all the classes and particularly relevant on cycle which is the most underrepresented class. That is, for large-scale object detection, the convergence is slower but, in the long term, adding the co-training outperforms the baseline.  

\begin{table}[!t]
\caption{\textbf{Object detection}. Comparison of our approach using different self-supervised tasks to the baseline. Models are trained during 500K steps.}
\label{tab:detection}
\begin{center}
\begin{tabular}{ c c c c c c}
\hline
 \multirow{2}{*}{Model} & self-sup. & \multicolumn{4}{c}{mAP (\%)} \\ 
 & task & Vehicle & Cycle & Person & Avg \\
 \hline
 baseline & -- & 91.12 & 71.41 & 75.6 & 79.38\\  
 Co-train & Jigsaw & 90.92 & 70.64 & 75.86 & 79.14\\
 Co-train & Rotation & \textbf{91.38} & \textbf{71.82} & \textbf{76.26} & \textbf{79.82}\\
\hline
\end{tabular}
\end{center}
\vspace{-0.5cm}
\end{table}

\subsection{Domain generalization and out-of-distribution robustness}
\begin{figure*}[!t]
  \centering
  \resizebox{0.94\textwidth}{!}{
  \begin{tabular}{cc|cc|cc}
  \small{day data} &\small{night and day data}&\small{day data} &\small{night and day data}&\small{day data} &\small{night and day data}\\
    \hspace{-0.15cm}\includegraphics[width=0.2\textwidth]{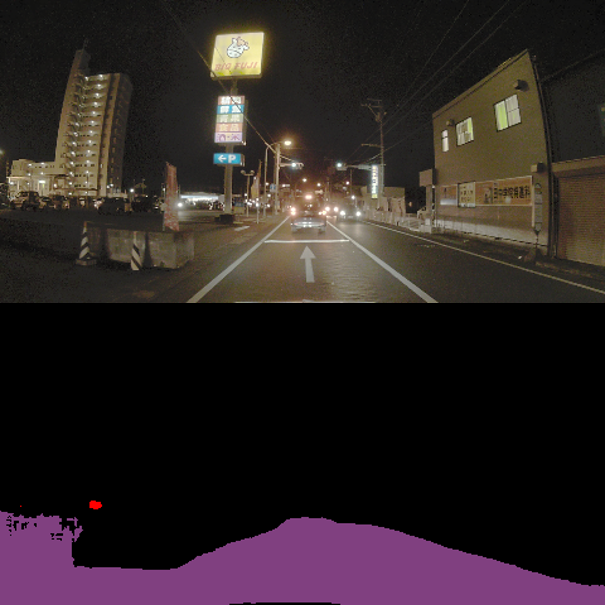}&
    \hspace{-0.350cm}\includegraphics[width=0.2\textwidth]{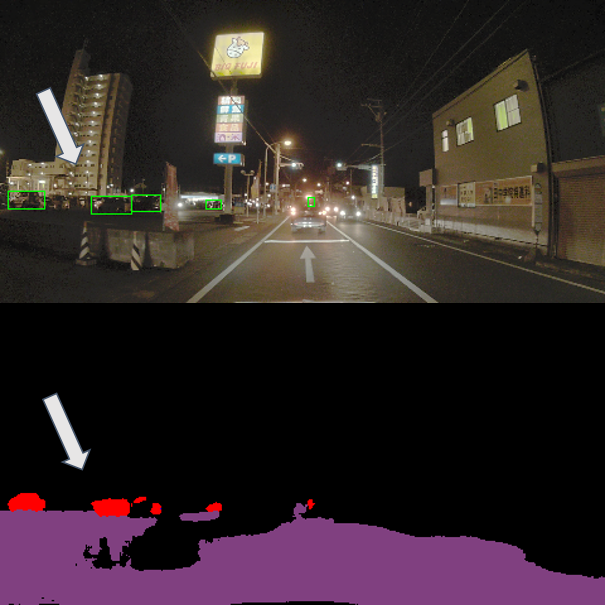}&
    \hspace{-0.15cm}\includegraphics[width=0.2\textwidth]{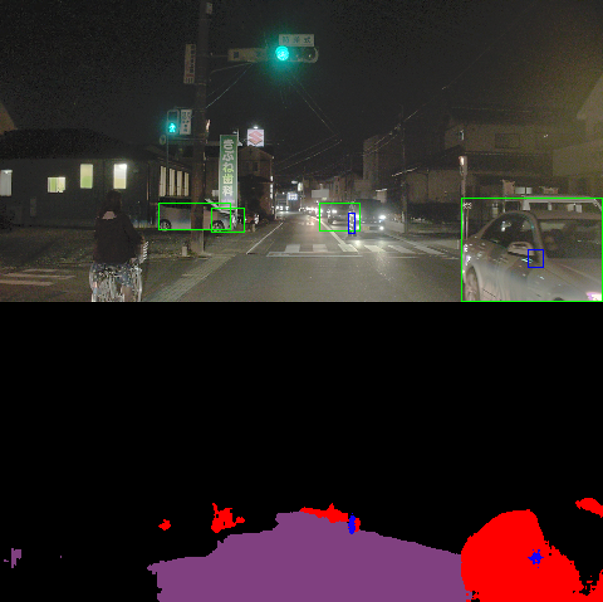}&
    \hspace{-0.35cm}\includegraphics[width=0.2\textwidth]{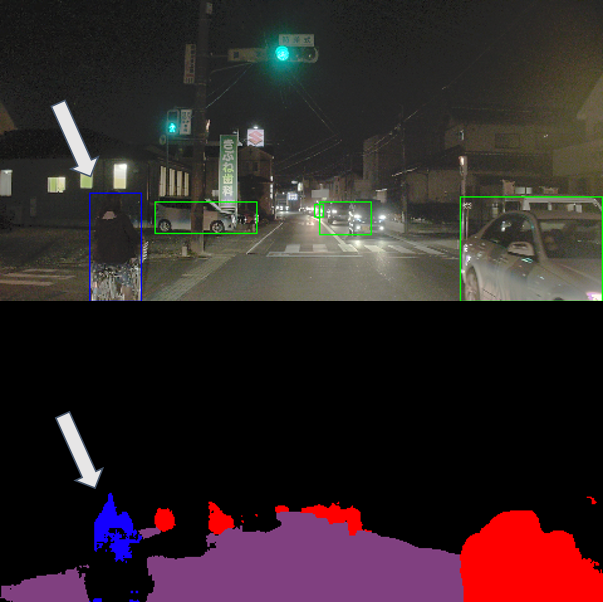}&
   \hspace{-0.15cm}\includegraphics[width=0.2\textwidth]{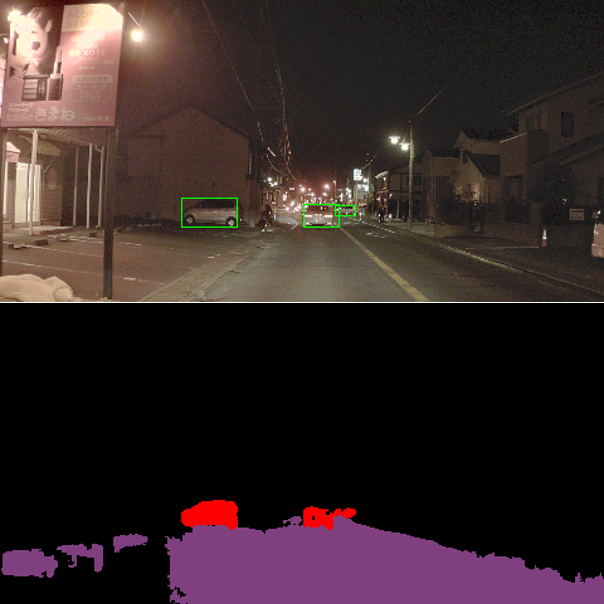}&
   \hspace{-0.350cm}\includegraphics[width=0.2\textwidth]{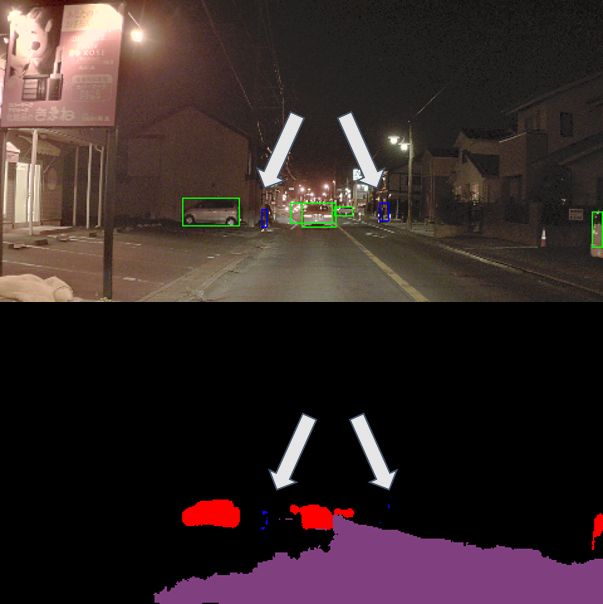}\\  
    \multicolumn{2}{c}{\hspace{-0.25cm}(a)}&\multicolumn{2}{c}{\hspace{-0.25cm}(b)}&\multicolumn{2}{c}{\hspace{-0.25cm}(c)}\\
  \end{tabular}
  }
  \vspace{-0.15cm}
  \caption{Qualitative results of the effect of adding off-distribution data to the self-supervised task. Left shows results using only day (unlabeled) data during training and, on the right, we show results using day and night data during the training of the self-supervised task. For each example, the top row shows results over the original image and the bottom row shows the masks for three main classes. As shown on the right column and highlighted with gray arrows, adding unlabeled data from the target distribution improves the performance helping to: a) detect cars in the dark in a parking lot and provide better segmentation of the road; b) detect a person on a bicycle and avoid a false positive of a person; c) improve detecting people far away from the camera in the dark and provides a better segmentation of the road. }
  \label{dg1}
  \vspace{-0.35cm}
\end{figure*}

In this experiment, we analyze the effect of self-supervised co-training under domain shifts and out-of-distribution samples. First, similar to~\cite{carlucci2019domain}, we evaluate the benefits of co-training under different distributions. To this end, we compared the performance of a model trained on a labeled source domain dataset to the performance of the same model co-trained using unlabeled data from the target domain. More precisely, in this experiment, we treat daylight images as source domain data and night images as target domain data. We filter out night data from the panoptic segmentation dataset and use it as source data. As we have an object detection night dataset, we directly use it as target domain data. Note that we don't use the labels provided by the object detection night training data but use its test data for evaluation. When evaluating on the object detection dataset, the prediction masks of objects from panoptic segmentation models are converted to bounding boxes. To show the effect of self-supervised co-training, we trained the panoptic segmentation model with the following methods:
\begin{enumerate}
  \item Baseline model that uses only panoptic segmentation daylight data.
  \item Self-supervised co-train where both supervised and self-supervised branches use panoptic segmentation daylight data.
  \item Self-supervised co-train where supervised branch uses panoptic segmentation daylight data and self-supervised branch uses panoptic segmentation daylight data + object detection night training data.
\end{enumerate}
Table~\ref{table:dg1} summarizes the quantitative results for this experiment. We observe that both jigsaw and rotation tasks help domain generalization when the target domain data is not present. This performance improvement is even larger when we use unlabeled target domain data for training the self-supervised branch. Fig.~\ref{dg1} shows representative qualitative results for this experiment. As shown, adding target domain data (night data) into the training of the self-supervised branch could improve the detection of small and dark objects, as well as providing a better segmentation. From these results, we can conclude that the model trained with unlabeled night data performs consistently better at identifying small and dark objects.

\begin{table}[!t]
\caption{\textbf{Object detection.} Robustness to domain shift. Co-training using unlabeled data improves the robustness of object detection to domain shifts.}
\label{table:dg1}
\vspace{-0.5cm}
\begin{center}
\resizebox{\columnwidth}{!}{
\begin{tabular}{c | c c | c c | c c | c c c | c}
\cline{2-11}
  & \multicolumn{2}{c|}{self-sup. task} & \multicolumn{2}{c|}{unlabeled data} & \multicolumn{2}{c|}{test data} & \multicolumn{3}{c|}{class}& global \\ 
&jigsaw&rotation& day & night & day & night & Vehicle & Cycle & Person & Avg \\\cline{1-11}
baseline & \multicolumn{2}{c|}{--}  & \checkmark & & & \checkmark &  45.86 & 35.29 & 47.63 & 42.93\\  \hline 
Co-train &  \checkmark & & \checkmark &  & & \checkmark &49.37 & 32.5 & 48.13 & 43.33\\
Co-train &  \checkmark & & \checkmark & \checkmark & & \checkmark & \textbf{50.76} & 36.25 & \textbf{48.45} & \textbf{45.15}\\ \hline 
Co-train &  & \checkmark & \checkmark &  & & \checkmark &50.71 & 35.76 & 47.58 & 44.68\\
Co-train &  & \checkmark & \checkmark & \checkmark & & \checkmark & 49.98 & \textbf{36.52} & 47.82 & 44.77\\
\hline
\end{tabular}
}
\end{center}
\vspace{-0.5cm}
\end{table}

\subsubsection{Out-of-distribution Robustness}
\begin{figure}[!t]
  \centering
  \resizebox{\columnwidth}{!}{
  \begin{tabular}{cccc}
       \hspace{-0.025cm}\includegraphics[width=0.25\columnwidth]{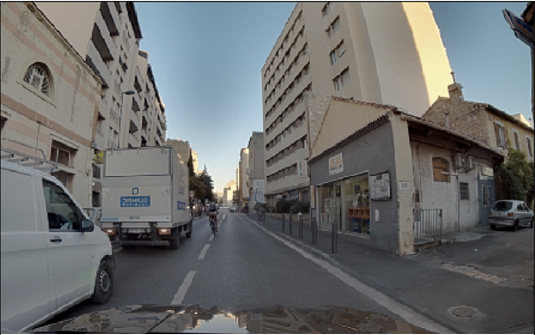}& 
       \hspace{-0.4cm}\includegraphics[width=0.25\columnwidth]{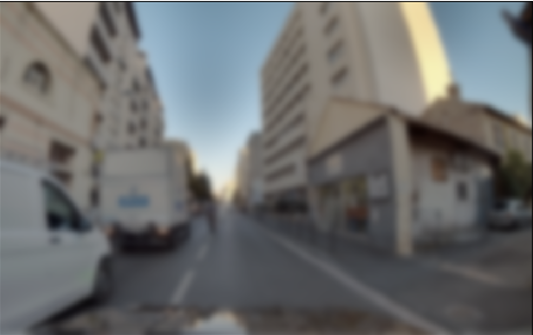}&
       \hspace{-0.4cm}\includegraphics[width=0.25\columnwidth]{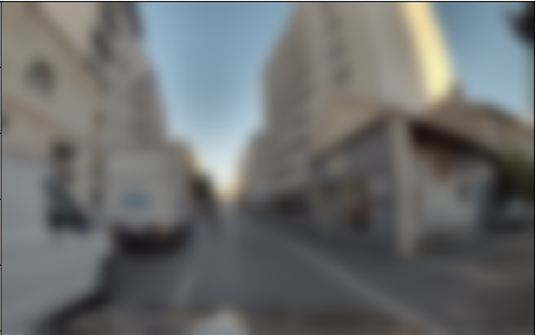}&
       \hspace{-0.4cm}\includegraphics[width=0.25\columnwidth]{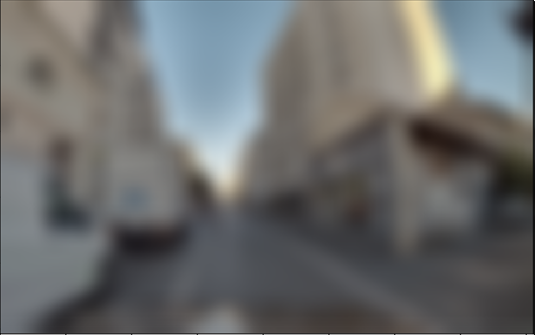}\\
       \small{original/no blur}&\small{$\sigma=5$}&\small{$\sigma=10$}&\small{$\sigma=15$}
  \end{tabular}
  }
  \caption{We use Gaussian noise to emulate out-of-distribution samples during test time. During training, Gaussian noise was not used as data augmentation.}
  \label{fig:gaussian}
  \vspace{-0.5cm}
\end{figure}

In a last experiment, we focus on testing the robustness of our approach to out-of-distribution samples. In the real world, out-of-distribution might occur due to complicated and unpredictable situations such as camera out of focus, rain falling on the camera, or camera malfunction, among many others. Here, we focus on evaluating our approach to noisy inputs. In particular, we compare the regression in performance of our approach to the baseline when the input is affected by Gaussian noise. We train both, the baseline and our approach, using the original, non-blurred data with data augmentation that does not include Gaussian noise. Then, we evaluate the performance of both models when test data is affected by different amount of Gaussian noise, see~\fig{fig:gaussian}. Table~\ref{table:dg3} summarizes the results for this experiments for object detection using both self-supervised tasks: jigsaw and rotation. As expected, the performance drop increases as the amount of noise in the input data increases. This performance degradation is also observed for our approaches independently of the self-supervised task. This degradation seems more severe for jigsaw compared to the baseline. Interestingly, using rotation as the self-supervised task yields slightly better results when images are severely blurred. 



\begin{table}[!t]
\caption{\textbf{Object detection}. mAP as a function of the input (Gaussian) noise.}
\vspace{-0.5cm}
\label{table:dg3}
\begin{center}
\begin{tabular}{ c c c c c c}
\hline
 Model & \multicolumn{1}{c}{self-sup. task}& No blur & $\sigma=5$ & $\sigma=10$ & $\sigma=15$ \\
 \hline
 baseline &\multicolumn{1}{c}{--}& $77.62$ &  $\mathbf{39.50}$ & $22.34$ & $15.36$ \\  
 Co-train & Jigsaw & $77.58$ & $38.81$ & $20.86$ & $13.99$ \\
 Co-train & Rotation &  $\mathbf{77.67}$ & $39.27$ &  $\mathbf{22.82}$ & $\mathbf{15.42}$ \\
\hline
\end{tabular}
\end{center}
\vspace{-0.5cm}
\end{table}

\section{Conclusions}

Deep Supervised learning is the main method to obtain state-of-the-art results in most computer vision tasks. Recently, there has been wide accessibility to acquiring data, however, annotation is still an expensive and time consuming process. In this paper, we described our system for improving existing production-based supervised perception tasks such as object detection by leveraging unsupervised data with minimum disruption to existing production pipelines. The main component of our system is a self-supervised branch added to main tasks in a multi-task framework. Our framework adds minimal change to the current supervised learning pipeline, and introduces minimal training cost and extra parameters. Its flexibility enables it to be easily extended to other self-supervised methods as well as other supervised tasks. We evaluated the contribution of two self-supervised tasks to two perception tasks: object detection and panoptic segmentation. Our experimental results show performance improvements of our approach over the baselines.

\bibliographystyle{IEEEtranBST/IEEEtran}
\bibliography{bib}

\begin{thebibliography}{10}
\providecommand{\url}[1]{#1}
\csname url@rmstyle\endcsname
\providecommand{\newblock}{\relax}
\providecommand{\bibinfo}[2]{#2}
\providecommand\BIBentrySTDinterwordspacing{\spaceskip=0pt\relax}
\providecommand\BIBentryALTinterwordstretchfactor{4}
\providecommand\BIBentryALTinterwordspacing{\spaceskip=\fontdimen2\font plus
\BIBentryALTinterwordstretchfactor\fontdimen3\font minus
  \fontdimen4\font\relax}
\providecommand\BIBforeignlanguage[2]{{%
\expandafter\ifx\csname l@#1\endcsname\relax
\typeout{** WARNING: IEEEtran.bst: No hyphenation pattern has been}%
\typeout{** loaded for the language `#1'. Using the pattern for}%
\typeout{** the default language instead.}%
\else
\language=\csname l@#1\endcsname
\fi
#2}}

\bibitem{gidaris2018unsupervised}
S.~Gidaris, P.~Singh, and N.~Komodakis, ``Unsupervised representation learning
  by predicting image rotations,'' \emph{arXiv preprint arXiv:1803.07728},
  2018.

\bibitem{noroozi2016unsupervised}
M.~Noroozi and P.~Favaro, ``Unsupervised learning of visual representations by
  solving jigsaw puzzles,'' in \emph{European conference on computer
  vision}.\hskip 1em plus 0.5em minus 0.4em\relax Springer, 2016, pp. 69--84.

\bibitem{radford2015unsupervised}
A.~Radford, L.~Metz, and S.~Chintala, ``Unsupervised representation learning
  with deep convolutional generative adversarial networks,'' \emph{arXiv
  preprint arXiv:1511.06434}, 2015.

\bibitem{donahue2016adversarial}
J.~Donahue, P.~Kr{\"a}henb{\"u}hl, and T.~Darrell, ``Adversarial feature
  learning,'' \emph{arXiv preprint arXiv:1605.09782}, 2016.

\bibitem{dosovitskiy2015discriminative}
A.~Dosovitskiy, P.~Fischer, J.~T. Springenberg, M.~Riedmiller, and T.~Brox,
  ``Discriminative unsupervised feature learning with exemplar convolutional
  neural networks,'' \emph{IEEE transactions on pattern analysis and machine
  intelligence}, vol.~38, no.~9, pp. 1734--1747, 2015.

\bibitem{carlucci2019domain}
F.~M. Carlucci, A.~D'Innocente, S.~Bucci, B.~Caputo, and T.~Tommasi, ``Domain
  generalization by solving jigsaw puzzles,'' in \emph{Proceedings of the
  IEEE/CVF Conference on Computer Vision and Pattern Recognition}, 2019, pp.
  2229--2238.

\bibitem{doersch2015unsupervised}
C.~Doersch, A.~Gupta, and A.~A. Efros, ``Unsupervised visual representation
  learning by context prediction,'' in \emph{Proceedings of the IEEE
  international conference on computer vision}, 2015, pp. 1422--1430.

\bibitem{zhang2016colorful}
R.~Zhang, P.~Isola, and A.~A. Efros, ``Colorful image colorization,'' in
  \emph{European conference on computer vision}.\hskip 1em plus 0.5em minus
  0.4em\relax Springer, 2016, pp. 649--666.

\bibitem{vincent2008extracting}
P.~Vincent, H.~Larochelle, Y.~Bengio, and P.-A. Manzagol, ``Extracting and
  composing robust features with denoising autoencoders,'' in \emph{Proceedings
  of the 25th international conference on Machine learning}, 2008, pp.
  1096--1103.

\bibitem{pathak2016context}
D.~Pathak, P.~Krahenbuhl, J.~Donahue, T.~Darrell, and A.~A. Efros, ``Context
  encoders: Feature learning by inpainting,'' in \emph{Proceedings of the IEEE
  conference on computer vision and pattern recognition}, 2016, pp. 2536--2544.

\bibitem{zhang2017split}
R.~Zhang, P.~Isola, and A.~A. Efros, ``Split-brain autoencoders: Unsupervised
  learning by cross-channel prediction,'' in \emph{Proceedings of the IEEE
  Conference on Computer Vision and Pattern Recognition}, 2017, pp. 1058--1067.

\bibitem{chen2020simple}
T.~Chen, S.~Kornblith, M.~Norouzi, and G.~Hinton, ``A simple framework for
  contrastive learning of visual representations,'' in \emph{International
  conference on machine learning}.\hskip 1em plus 0.5em minus 0.4em\relax PMLR,
  2020, pp. 1597--1607.

\bibitem{misra2020self}
I.~Misra and L.~v.~d. Maaten, ``Self-supervised learning of pretext-invariant
  representations,'' in \emph{Proceedings of the IEEE/CVF Conference on
  Computer Vision and Pattern Recognition}, 2020, pp. 6707--6717.

\bibitem{caruana1997multitask}
R.~Caruana, ``Multitask learning,'' \emph{Machine learning}, vol.~28, no.~1,
  pp. 41--75, 1997.

\bibitem{thrun1998learning}
S.~Thrun and L.~Pratt, ``Learning to learn: Introduction and overview,'' in
  \emph{Learning to learn}.\hskip 1em plus 0.5em minus 0.4em\relax Springer,
  1998, pp. 3--17.

\bibitem{bilen2016integrated}
H.~Bilen and A.~Vedaldi, ``Integrated perception with recurrent multi-task
  neural networks,'' \emph{arXiv preprint arXiv:1606.01735}, 2016.

\bibitem{misra2016cross}
I.~Misra, A.~Shrivastava, A.~Gupta, and M.~Hebert, ``Cross-stitch networks for
  multi-task learning,'' in \emph{Proceedings of the IEEE conference on
  computer vision and pattern recognition}, 2016, pp. 3994--4003.

\bibitem{long2015learning}
M.~Long and J.~Wang, ``Learning multiple tasks with deep relationship
  networks,'' \emph{arXiv preprint arXiv:1506.02117}, vol.~2, no.~1, 2015.

\bibitem{lu2017fully}
Y.~Lu, A.~Kumar, S.~Zhai, Y.~Cheng, T.~Javidi, and R.~Feris, ``Fully-adaptive
  feature sharing in multi-task networks with applications in person attribute
  classification,'' in \emph{Proceedings of the IEEE conference on computer
  vision and pattern recognition}, 2017, pp. 5334--5343.

\bibitem{yang2016trace}
Y.~Yang and T.~M. Hospedales, ``Trace norm regularised deep multi-task
  learning,'' \emph{arXiv preprint arXiv:1606.04038}, 2016.

\bibitem{sener2018multi}
O.~Sener and V.~Koltun, ``Multi-task learning as multi-objective
  optimization,'' \emph{arXiv preprint arXiv:1810.04650}, 2018.

\bibitem{ren2018cross}
Z.~Ren and Y.~J. Lee, ``Cross-domain self-supervised multi-task feature
  learning using synthetic imagery,'' in \emph{Proceedings of the IEEE
  Conference on Computer Vision and Pattern Recognition}, 2018, pp. 762--771.

\bibitem{doersch2017multi}
C.~Doersch and A.~Zisserman, ``Multi-task self-supervised visual learning,'' in
  \emph{Proceedings of the IEEE International Conference on Computer Vision},
  2017, pp. 2051--2060.

\bibitem{han2020self}
T.~Han, W.~Xie, and A.~Zisserman, ``Self-supervised co-training for video
  representation learning,'' \emph{arXiv preprint arXiv:2010.09709}, 2020.

\bibitem{kundu2019adapt}
J.~N. Kundu, N.~Lakkakula, and R.~V. Babu, ``Um-adapt: Unsupervised multi-task
  adaptation using adversarial cross-task distillation,'' in \emph{Proceedings
  of the IEEE/CVF International Conference on Computer Vision}, 2019, pp.
  1436--1445.

\bibitem{klingner2020improved}
M.~Klingner, A.~Bar, and T.~Fingscheidt, ``Improved noise and attack robustness
  for semantic segmentation by using multi-task training with self-supervised
  depth estimation,'' in \emph{Proceedings of the IEEE/CVF Conference on
  Computer Vision and Pattern Recognition Workshops}, 2020, pp. 320--321.

\bibitem{lee2019multi}
W.~Lee, J.~Na, and G.~Kim, ``Multi-task self-supervised object detection via
  recycling of bounding box annotations,'' in \emph{Proceedings of the IEEE/CVF
  Conference on Computer Vision and Pattern Recognition}, 2019, pp. 4984--4993.

\bibitem{kirillov2019panoptic}
A.~Kirillov, K.~He, R.~Girshick, C.~Rother, and P.~Doll{\'a}r, ``Panoptic
  segmentation,'' in \emph{Proceedings of the IEEE/CVF Conference on Computer
  Vision and Pattern Recognition}, 2019, pp. 9404--9413.

\end{thebibliography}
\end{document}